\documentclass[11pt,a4paper]{article}
\usepackage[hyperref]{acl2020}
\usepackage{times}
\usepackage{latexsym}
\usepackage{graphicx}

\usepackage{microtype}

\aclfinalcopy 
\def\aclpaperid{29} 

\title{Identifying Hijacked Reviews}

\author{Monika Daryani \\
  Texas A\&M University \\
  College Station, TX 77843 \\
  \texttt{monikadaryani@tamu.edu} \\
  \And
  James Caverlee \\
  Texas A\&M University \\
  College Station, TX 77843 \\
  \texttt{caverlee@tamu.edu} \\}

\date{}

\begin{document}
\maketitle
\begin{abstract}

Fake reviews and review manipulation are growing problems on online marketplaces globally. \textit{Review Hijacking} is a new review manipulation tactic in which unethical sellers ``hijack'' an existing product page (usually one with many positive reviews), then update the product details like title, photo, and description with those of an entirely different product. With the earlier reviews still attached, the new item appears well-reviewed. However, there are no public datasets of review hijacking and little is known in the literature about this tactic. Hence, this paper proposes a three-part study: (i) we propose a framework to generate synthetically labeled data for review hijacking by swapping products and reviews; (ii) then, we evaluate the potential of both a Twin LSTM network and BERT sequence pair classifier to distinguish legitimate reviews from hijacked ones using this data; and (iii) we then deploy the best performing model on a collection of 31K products (with 6.5 M reviews) in the original data, where we find 100s of previously unknown examples of review hijacking.
\end{abstract}

\section{Introduction}
Reviews are an essential component of many online marketplaces, helping new consumers assess product quality, legitimacy, and reliability. Recent surveys indicate that an overwhelming majority of people read reviews \cite{consumerreview}. Indeed, 79\% of people overall and 91\% of people ages 18-34 trust online reviews as much as personal recommendations \cite{productreview}. Naturally, reviews have become a target of manipulation, misuse, and abuse \cite{mukherjee2012spotting}. 

In this paper, we focus on the problem of \textit{review hijacking}, a relatively new attack vector and one that has received little, if any, research attention. Review hijacking is a fraud technique wherein a blackhat seller ``hijacks'' a product page that typically has already accumulated many positive reviews and then replaces the hijacked product with a different product (typically one without any positive reviews). The sellers then reap the ratings ``halo'' from consumers who assume the new product is highly rated. This review hijacking (also referred to as \textit{review reuse} or \textit{bait-and-switch reviews}) provides the sellers with a shortcut to many undeserved positive reviews.

\begin{figure}[h]
	\centering
	\includegraphics[width=1.0\linewidth]{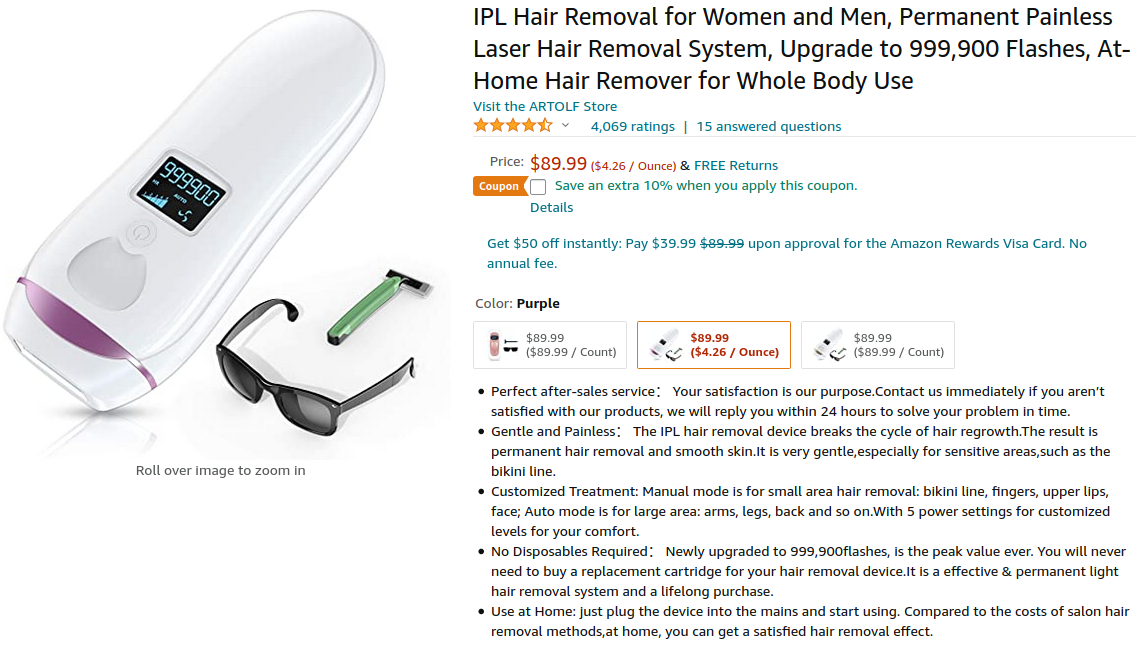}
	\caption{An example of review hijacking on Amazon (May 7, 2021)}
	\label{fig:hairremover}
\end{figure}

\begin{figure}[h]
	\centering
	\includegraphics[width=1.0\linewidth]{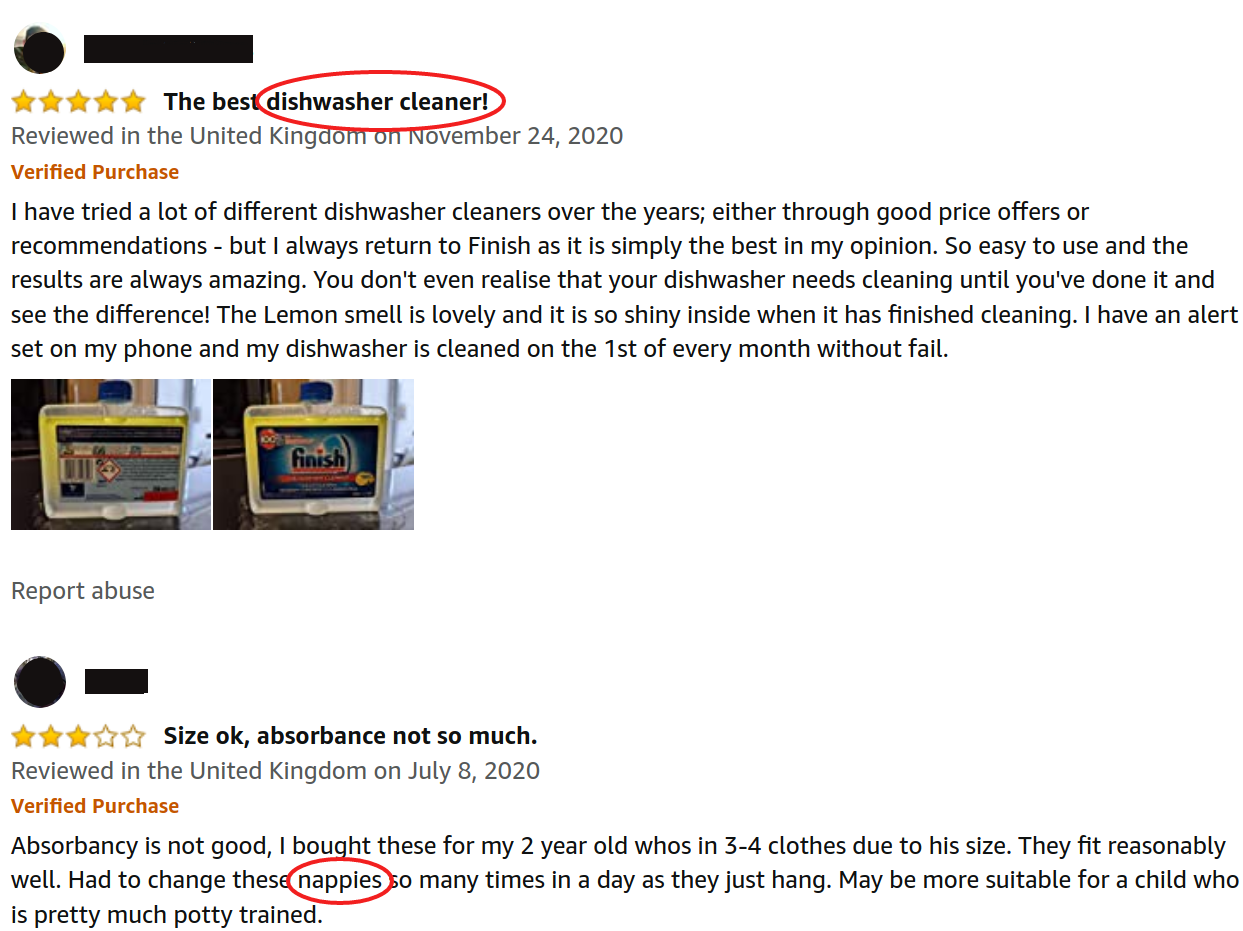}
	\caption{Hijacked reviews associated with the hair removal product in Figure~\ref{fig:hairremover}}
	\label{fig:hairremoverreviews}
\end{figure}

An example is shown in Figure~\ref{fig:hairremover} which we discovered in the first week of May 2021. This hair removal product has 4,069 reviews with an average rating of close to five stars. On inspection of the reviews (see Figure~\ref{fig:hairremoverreviews}), we find many that refer to other products like dishwasher cleaners and diapers. Also, these reviews are for verified purchases which can provide added weight to the ostensible veracity of the reviews. 

We have identified at least three different methods that blackhat sellers adopt to conduct review hijacking depending on the particular e-commerce platform. A seller can incrementally change aspects of their own product (like the title, photo, description), resulting in an entirely new product, though still associated with the original reviews. Alternatively, a seller can add his product as a product variation of some other product to aggregate reviews from the former product. One can also merge reviews from some other products to their own by changing country or using some other platform-specific loopholes.

While review hijacking has been recognized in the press and social media as a growing problem, e.g., \cite{CONREP2, ebayreviewabuse, reviewfraud, stackexchangeswapped, buzzfeed, theverge}, there has been no structured research to date on identifying review hijacking. We attribute this to several key challenges: 

\begin{itemize}
\item First, there are no standard datasets of review hijacking, nor are there gold labels of known examples. Hence, it is challenging to validate models that aim to uncover review hijacking. 
\item Second, review hijacking is a targeted attack vector with a skewed distribution, and so there are no simple approaches to find examples. In a preliminary investigation, we manually labeled hundreds of reviews and found fewer than 0.01\% reviews that could be considered part of a review hijacking attack.
\item Third, many reviews cannot easily be labeled as hijacked or not. For example, reviews like ``Great product! Five stars!'' are generic and could potentially be associated with any product.
\item Finally, hijackers may adopt sophisticated techniques to avoid detection. For example, some products may have a mix of legitimate reviews to camouflage the hijacked ones (e.g., by incentivizing reviewers to contribute a review about the hair removal product). 
\end{itemize}

Hence, this paper proposes an initial investigation into the potential of identifying review hijacking. We conduct a three-part study. Due to the challenges of finding high-quality examples of review hijacking, we first propose a framework to generate synthetic examples of review hijacking by swapping products and reviews. We do so both at the inter-category level (where presumably it should be easier to determine if a review is associated with a product) and at the intra-category level (where product similarity within the category may make this more challenging). Over this synthetic dataset, we evaluate the potential of both a Twin LSTM network and BERT sequence pair classifier to distinguish legitimate reviews from hijacked ones. Based on the encouraging results from this experiment, we then deploy the BERT sequence pair classifier algorithm on a real collection of 31K products (with 6.5 M reviews). By averaging the review scores from the classifier for each product, we find that products with an average review score (or  \textit{suspiciousness} score) $> 0.5$ have $99.95\%$ of the listings containing unrelated or hijacked reviews. These findings suggest the promise of large-scale detection of review hijacking in the wild.

\section{Related Work}
The manipulation of reviews and review platforms has been widely studied, e.g.,  \cite{gossling2018manager, jindal2007, kaghazgaran2017behavioral, mukherjee2012spotting, mukherjee2013fake}, though there is little research literature on the problem of review hijacking. Here, we highlight several efforts related to the methods proposed in this paper. Higgins et al. developed models for an essay rating system to detect bad-faith essays by comparing the essay titles to the essay text to determine whether the title and text were in agreement through the use of word similarity \cite{higgins2006identifying}. A similar idea motivates our approach that compares product titles/descriptions with review text. Louis and Higgins continued this line of research to determine whether a particular essay was related to the essay prompt or question by expanding short prompts and spell correcting the texts \cite{10.5555/1866795.1866808}. Rei and Cummins extended this work and combined various sentence similarity measures like TF-IDF and Word2Vec embeddings with moderate improvement over Higgins' work \cite{rei-cummins-2016-sentence}. Apart from the essay space, Ryu et al. investigated the detection of out-of-domain sentences \cite{RYU201726}. They proposed a neural sentence embedding method representing sentences in a low-dimensional continuous vector space that emphasizes aspects in-domain and out-of-domain for a given scenario. In another direction, fake news detection and clickbait detection can be viewed as related tasks. For example, Hanselowski et al.used a BiLSTM model with attention to determine if the headline of a news article agrees, disagrees, or is unrelated to the text as part of a Fake News Challenge \cite{hanselowski-etal-2018-retrospective}.

\section{Generating Synthetic Examples of Review Hijacking}
In our preliminary investigation, we examined hundreds of reviews from the Amazon dataset provided by McAuley \cite{ni2019justifying}. The dataset contains 233.1 million reviews from May 1996 to October 2018, with reviews and product information including title, description, etc. However, we find very few examples of review hijacking. Hence, we concluded that hiring crowd labelers or subject matter experts to label product-review pairs as hijacked or not hijacked might not be fruitful. Instead, we propose a method to generate synthetic examples for studying the potential of models to identify hijacked reviews. 

\subsection{Preliminaries}
As a first step, we prepared the Amazon dataset. For each product $i$, we combined the description  (product text provided by the seller), title (the name of the product), the brand of the product, and features (product features like color or size) into a single \textit{product text} $P_i$. We also removed products with fewer than five reviews.

For each review $j$, we combined the reviewText  (the text in the review body), the style (which contains some optional product features like color or size), and summary (which is the headline of the review) into a single \textit{review text} $R_j$.

Hence, our goal is to determine if each review $j$ associated with the product $i$, is actually related to the product or not. If the review is unrelated, we can conclude that there is potential evidence of review hijacking for the product. Of course, there could be other reasons for a review for being unrelated to a product, like an error by the reviewer. We leave this fine-grained determination as future work.

\subsection{Swapping Reviews}\label{sec:swapreviews}
Given these products and reviews, we propose to randomly swap reviews between a pair of distinct products, yielding a collection of unrelated product-review pairs. As a first step, we assume that all reviews are actually related to the associated product. Hence, we have a large set of product-review pairs with the label \textbf{related} (= 0). Of course, we know that our data has some hijacked reviews (on the order of $< 0.01 \% $), so we will tolerate some errors in these labels.

\begin{figure}[!ht]
	\centering
	\includegraphics[width=1.0\linewidth]{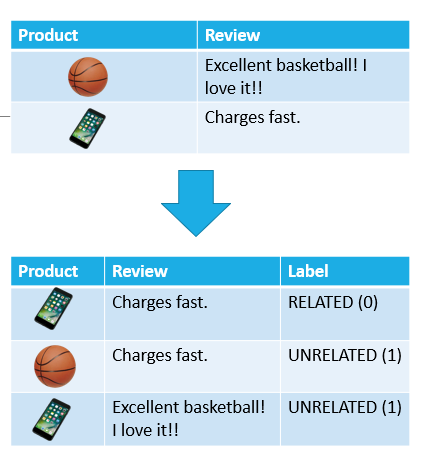}
	\caption{Generation of Synthetic Label and Data by Swapping Reviews among Dissimilar Products}
	\label{fig:label_gen}
\end{figure}

By randomly swapping product-review pairs, we get a set of product-review pairs with the label \textbf{unrelated} (= 1). For example, Figure \ref{fig:label_gen} shows a simple example of a basketball and a phone, each with an associated review. We swap reviews among the products to generate \textbf{unrelated} (= 1) labels in addition to the original \textbf{related} (= 0) labels.


But, how do we select which products to select for randomly swapping reviews? Randomly selecting products may lead to such a clear mismatch between the review text and the product text that detection would be trivial. On the other hand, selecting closely related products (e.g., by selecting Samsung mobile covers from two different brands) may yield reviews that are essentially undetectable as possible hijacking. Hence, we propose two methods for finding pairs of dissimilar products for review swapping.


\medskip
\noindent\textbf{Inter-Category Swapping.} The first approach takes a product text $P_i$ composed of the title, features, and description from one category (e.g., Beauty, Clothing, Electronics) and the review texts $R_j$ of a product in another category for \textbf{unrelated} reviews. For \textbf{related} reviews, we take the original product-review pairs. We obtained a set of $\approx 59k$  reviews with $\approx 25k$ unrelated reviews and $\approx 34k$ related reviews. 

\medskip
\noindent\textbf{Intra-category Swapping.} The first approach handles hijacking across categories. For hijacking occurring within a product category, we use Jaccard distance. We converted product titles for each product into TF-IDF feature matrices, found pairwise Jaccard distances between them, and we formed product pairs ($A_1, A_2$) with Jaccard distance 0. Then, we took the product text $P_i$ of one product $A_1$, and the review text $R_j$ of another product $A_2$ and labeled this as \textbf{unrelated}. Similarly, we took the product text of $A_2$ and a review of $A_1$ as \textbf{unrelated}. For \textbf{related} labels, we took the product text, and the review text of $A_1$, and likewise for $A_2$ to get another set of related data. We obtained a set of $\approx 56k$ reviews with $\approx 22k$ unrelated reviews and $\approx 34k $ related reviews. 


\section{Identifying Synthetic Examples}
Given these synthetic datasets of hijacked reviews, can we detect them? In this section, we report on experiments with two approaches: one based on a Twin LSTM and one based on BERT Sentence Pair Classification. 

We shuffled the product-review pairs and split them into training, validation and test set in ratio 70:10:20 for both of the datasets. The actual number of reviews in each set depends on the swapping categories and is discussed in Section \ref{sec:swapreviews} We train on the train set, tune models on the validation set, and have reported results on the test set.



\subsection{Twin LSTM Network}
\begin{figure}[!ht]
	\centering
	\includegraphics[width=1.0\linewidth]{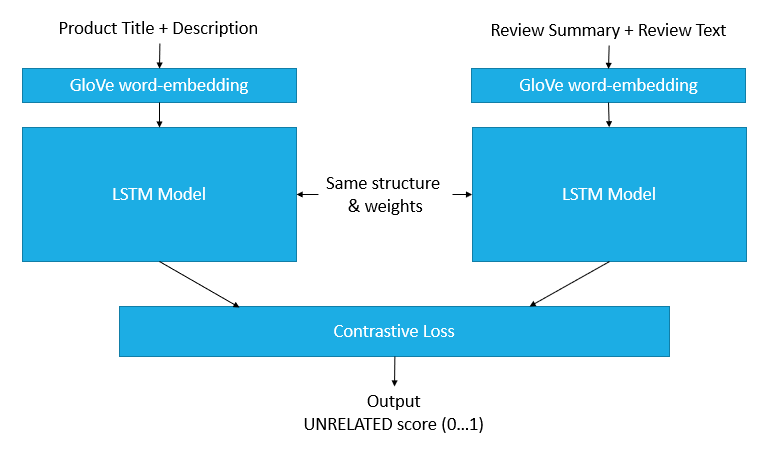}
	\caption{Twin LSTM Network}
	\label{fig:siamese}
\end{figure}

The first approach adopts a Twin neural network which has shown success in comparing images and text. This network uses the same weights in parallel in tandem on two inputs to return output based on the relation or distance between them \cite{Chicco2021}. Concretely, we compare sentence pairs and determine if they are similar or not. We tokenized our inputs and converted them into sequences. Then we used 300-dimensional GloVe \cite{pennington2014glove} embeddings and formed an embedding matrix for our tokens. We get two embedding matrices for both inputs, which we feed into the LSTM network illustrated in Figure \ref{fig:siamese}. We use twin LSTM networks with two layers of 64 nodes each, with a dropout of 0.01. We calculate the cosine similarity between the two input embeddings and evaluate the performance by computing cross-entropy loss using accuracy and AUC (Area Under Curve). It takes 13 epochs with Adam optimizer and learning rate of 0.00001 to get the result.

\subsection{BERT Sequence Pair Classifier}
\begin{figure}[!ht]
	\centering
	\includegraphics[width=1.0\linewidth]{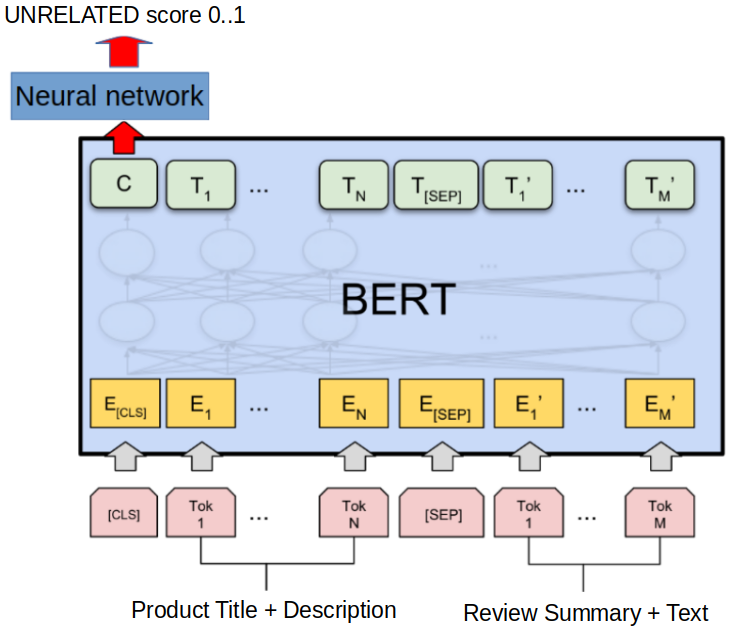}
	\caption{BERT Sequence Pair Classifier}
	\label{fig:bert}
\end{figure}

The second approach adopts the popular BERT pre-trained language model \cite{devlin-etal-2019-bert}. Since BERT provides a deep bidirectional representation, conditioned on text in both directions, we expect this method to perform better than the twin neural network, which uses GLOVE embeddings. Our model is prepared from the BERT BASE model (bert\_12\_768\_12) from \href{https://nlp.gluon.ai/v0.9.x/examples/sentence_embedding/bert.html}{GluonNLP}. We add a layer on top for classification, as shown in Figure \ref{fig:bert}. We use Adam optimizer for optimizing this classification layer and get results with only 3 epochs.

Now we form the sentence pairs for classification. Like the previous method, the first sentence is the product text $P_i$ (a concatenation of product title, features, and description). The second sentence is the review text $R_j$ (a concatenation of the review summary and review text). We then tokenize the sentences, insert [CLS] at the start, insert [SEP] at the end and between both the sentences, and generate segment ids to specify if a token belongs to the first sentence or the second one. We now run the BERT fine-tuning with these sequences as inputs. We get the output as an \textit{unrelated score} $u(i,j)$ between 0 and 1. For texts longer than 512 tokens, we truncate and take the first 512 tokens for our model.  As 99\% of the review texts have fewer than 512 tokens, this choice impacts very few reviews.




\subsection{Results}
Table \ref{tab:results_sup} shows the results reported on test data using these two approaches. We see that the Twin LSTM Network provides more than 80\% accuracy and high ROC result on both inter-category and intra-category datasets.  The BERT-based classifier has more than 90\% accuracy and ROC result for both datasets. We see that both methods perform better on the inter-category dataset than the intra-category one. In the inter-category dataset, we obtain unrelated reviews by taking products from one category and review texts from another. Hence, models trained on this dataset can learn product features of one category at a time and develop expertise in that category. The intra-category dataset is more challenging for both approaches. Since products are drawn from the same category, there can be less clarity in distinguishing features of the reviews.

\begin{figure}
	\centering
	\includegraphics[width=0.965\linewidth]{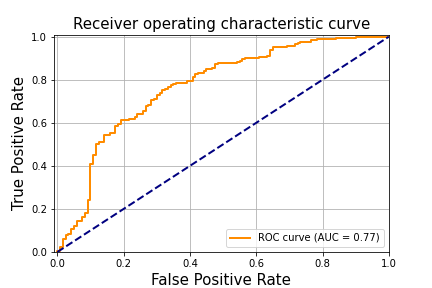}
	\caption{ROC curve for Twin LSTM network run on Intra-category data (Jaccard distance)}
	\label{fig:lstm_roc_pr_j}
	\includegraphics[width=0.965\linewidth]{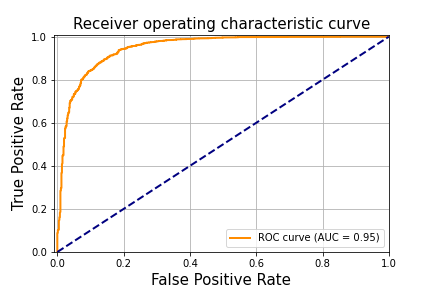}
	\caption{ROC curve for BERT seq. pair classifier run on Intra-category data (Jaccard distance)}
	\label{fig:bert_roc_pr_j}
	\includegraphics[width=0.965\linewidth]{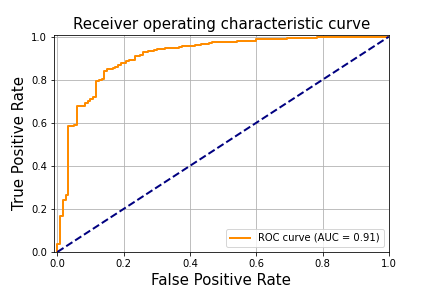}
	\caption{ROC curve for Twin LSTM network run on Inter-category data}
	\label{fig:lstm_roc_pr_ic}
	\includegraphics[width=0.965\linewidth]{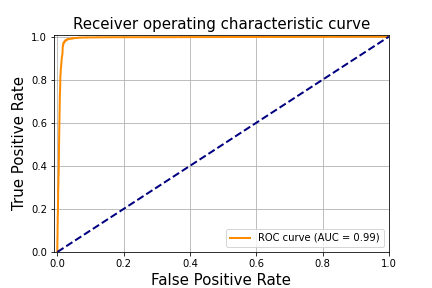}
	\caption{ROC curve for BERT seq. pair classifier run on Inter-category data}
	\label{fig:bert_roc_pr_ic}
\end{figure}

Paired with this summary table (Table \ref{tab:results_sup}), we show in  Figures~\ref{fig:lstm_roc_pr_j}, \ref{fig:bert_roc_pr_j}, \ref{fig:lstm_roc_pr_ic} and \ref{fig:bert_roc_pr_ic} the ROC curve for the BERT-based model and Twin LSTM network. We can clearly see that BERT-based model performs better than LSTM. We can also see how both models perform better on the inter-category dataset rather than the intra-category one. 

\begin{table*}[!ht]
	\centering
	\begin{tabular}{llll}
		\hline
		\textbf{Model} & \textbf{Synthetic Data} & \textbf{Accuracy} & \textbf{ROC result}\\
		\hline
		Twin LSTM Network & Intra-category & 0.823 & 0.770 \\
		 & Inter-category &  0.885 & 0.910 \\
		BERT Sequence Pair Classifier & Intra-category & 0.916 & 0.948  \\
		 & Inter-category & 0.965 & 0.993\\
		\hline
	\end{tabular}
	\caption{\label{tab:results_sup}
		Hijacked review detection accuracy reported on the test set
	}
\end{table*}



\section{Detecting Hijacked Reviews In-the-Wild}
Even though encouraging, these results are on synthetic data, and the data itself may contain noisy labels. Hence, we next turn to the task of uncovering hijacked reviews in-the-wild. Can a model trained over synthetic data identify actual hijacked reviews? 

\begin{figure*}[!ht]
	\centering
	\includegraphics[width=1.0\linewidth]{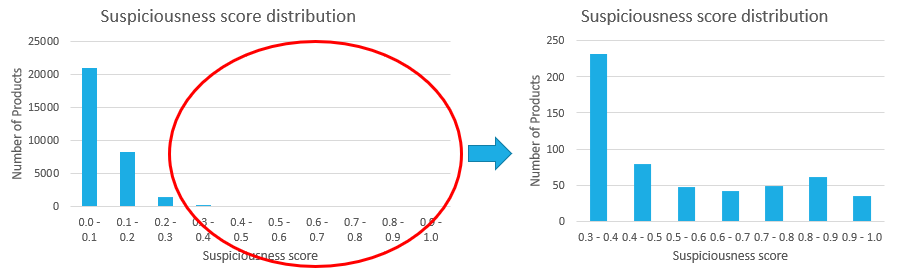}
	\caption{Average Review Score vs. Number of Products}
	\label{fig:eval}
\end{figure*}

\subsection{Approach and Results}
For this experiment, we used the BERT sequence pair classifier and applied it to a dataset of 31K products (with 6.5 M reviews) with the original product-review pairs intact. These 31K products were held out and not used during the training. 

For each product-review pair, we take the unrelated score output from the trained BERT-based model as $u(i,j)$. For a product $i$ with $n$ reviews, we calculate an average suspiciousness review score as follows:

$$
score_i = \frac{\sum_{j=1}^{n}\ u(i,j)}{n} 
$$

Based on this suspiciousness score, we plot the distribution of all 31K products in Figure~\ref{fig:eval}. Unsurprisingly, the vast majority of products have a very low suspiciousness score. About $99\%$ of products have scored $ < 0.3$, reinforcing our initial assumption about a skewed class distribution. In other words, the vast majority of the reviews on listings seem to be related to the product itself. However, we find many cases of potential review hijacking (see the right side of Figure~\ref{fig:eval}), indicating that this targeted attack is indeed a threat to review platforms. 

We manually checked a sample of 200+ products with a suspiciousness score of $> 0.5$. We found that all but one of the products contained reviews referring to a different product. While there is uncertainty as to the mechanism leading to an unrelated review, we hypothesize that these are indeed previously unknown cases of review hijacking. And in an encouraging direction, these results indicate the promise of training models over synthetic hijacked reviews for uncovering actual instances.



\subsection{Case Study}
In this section, we discuss three sample products three sample products and their distribution of unrelated scores $u(i,j)$ that are assigned by the BERT-based model. These three products are from the Cellphone \& Accessories category.

\begin{figure}[!ht]
	\centering
	\includegraphics[width=1.0\linewidth]{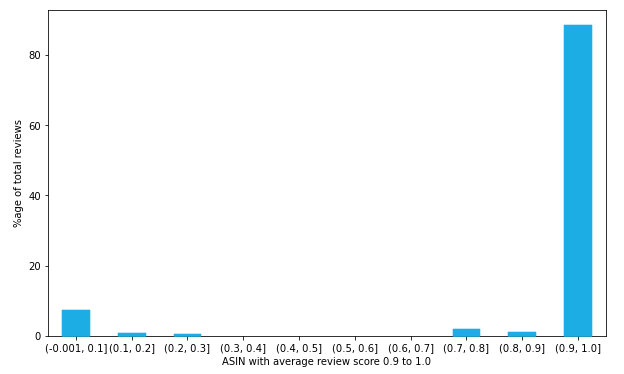}
	\caption{Unrelated Review Score Distribution for Product-1 showing predominantly unrelated reviews}
	\label{fig:sample_asin1}
\end{figure}

Figure \ref{fig:sample_asin1} shows the unrelated score distribution for all of the reviews of product-1. Product-1 has an average unrelated review score of $0.9$ to $1.0$. We can see from the distribution that most reviews have a high unrelated score $(> 0.9)$. We manually inspect these reviews and observe that these reviews are indeed unrelated. Hence, we conclude that this product is an example of review hijacking.

\begin{figure}[!ht]
	\centering
	\includegraphics[width=1.0\linewidth]{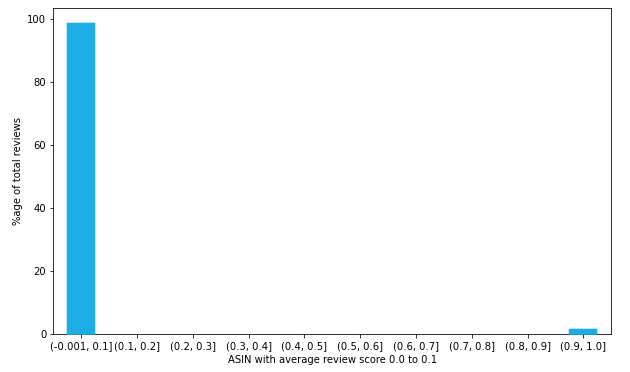}
	\caption{Unrelated Review Score Distribution for Product-2 showing predominantly related reviews}
	\label{fig:sample_asin2}
\end{figure}

Figure \ref{fig:sample_asin2} shows the unrelated score distribution for product-2. Product-2 has an average review score of $0.0$ to $0.1$, meaning most of the reviews seem appropriate. We can see from the distribution that most reviews have a low unrelated score $(< 0.1)$, and a few have a high score $(> 0.9)$. We manually inspect the reviews with high unrelated scores $(> 0.9)$ and observe that these reviews are either misclassified by our BERT-based model or do not have enough information to determine the label (e.g., reviews like ``Great Product!''). Thus, we conclude that this product is not an example of review hijacking.

\begin{figure}[!ht]
	\centering
	\includegraphics[width=1.0\linewidth]{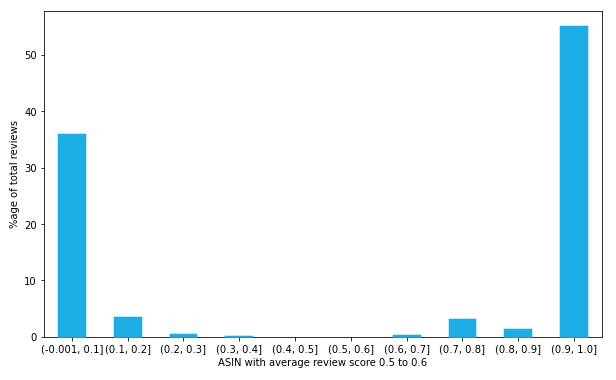}
	\caption{Unrelated Review Score Distribution for Product-3 showing a roughly equal mix of related and unrelated reviews}
	\label{fig:sample_asin3}
\end{figure}

Finally, Figure \ref{fig:sample_asin3} shows the unrelated review score distribution for product-3. Product-3 has an average review score of $0.5$ to $0.6$. We can see from the distribution that about $55\%$ of the reviews have a high unrelated score, while $35\%$ reviews have a low unrelated score. We manually inspect reviews with high unrelated scores $(> 0.9)$ and observe that these reviews are indeed unrelated to the product. We also inspect the reviews with low unrelated scores ($< 0.1$ and $< 0.2$) and observe that most are related to the product. As this product has a mix of related and unrelated reviews, we hypothesize that it is also an example of review hijacking containing some related reviews.


\section{Conclusion, Limitations and Next Steps}
This paper has examined the challenge of identifying hijacked reviews. Since we know little about these hijacked reviews, we first proposed to generate synthetic examples by swapping the reviews of a product with reviews on an unrelated product. We then tested the viability of a Twin LSTM network and BERT sentence pair classifier to uncover these unrelated reviews. Both approaches provided excellent results on synthetic data, but do they actually identify hijacked reviews in the wild? Our preliminary investigation showed that a model trained over synthetic data could detect many examples of previously unknown cases of review hijacking. 


Our method also has some limitations. First, the major drawback occurs because the data is labeled synthetically. Hence, there is no way to find the actual recall for our approach. Calculating recall requires manual labeling of all product-review pairs, which is an expensive process. Second, our method is dependent on the accuracy of labeling methods. For the intra-category case, our method cannot detect products hijacked with similar wording in the same category since their Jaccard distance is low. For example, if there are two products, ``iPhone X'' and ``iPhone 5C cover'', the products will have a low Jaccard distance, and the reviews hijacked among them cannot be labeled correctly. Therefore, our ML model can also not learn this kind of review hijacking. Third, generic reviews like ``Good product!'' and ``Product shipped fast'' were labeled hijacked and not hijacked depending on what product they belonged to. Ideally, we would want to label all of them as not hijacked. This random labeling adds to the noise in the labels.

In our continuing work, we are interested in two main directions: data and methods. From a data perspective, we are investigating more refined methods to generate synthetic labels. Can we couple crowd labelers with our swapping approach to construct better product-review pairs? We are also interested in updating the data itself. Our dataset covers reviews up to 2018, though many media reports of review hijacking were not until 2019. There could have been a rise in review hijacking that is not as prominent in our data. From a methods perspective, we have focused purely on text-based signals. Incorporating image-based features like from the product itself and user-submitted images could help identify examples of review hijacking. We are also interested in adopting recent advances in pre-trained language models like T5, DeBERTa, and RoBERTa. We are also focusing on using e-commerce specific text (like product catalog data) to instill domain-specific knowledge during the pre-training of language models versus BooksCorpus and English Wikipedia used in BERT.







\bibliography{acl2020}
\bibliographystyle{acl_natbib}

\end{document}